\documentclass[a4paper]{cas-sc}

\usepackage[numbers]{natbib}
\usepackage{booktabs}  
\usepackage{tabularx}  
\usepackage{threeparttable}
\usepackage{hyperref}
\def\tsc#1{\csdef{#1}{\textsc{\lowercase{#1}}\xspace}}
\tsc{WGM}
\tsc{QE}
\tsc{EP}
\tsc{PMS}
\tsc{BEC}
\tsc{DE}

\begin{document}
\let\WriteBookmarks\relax
\shorttitle{}
\shortauthors{S. Lin et~al.}

\makeatletter
\def\printorcid{}
\makeatother

\ExplSyntaxOn
\bool_gset_true:N \g_stm_nologo_bool
\ExplSyntaxOff

\title [mode = title]{Precision Synthesis of Multi-Tracer PET via VLM-Modulated Rectified Flow for Stratifying Mild Cognitive Impairment}   

\author[aff1,aff4]{Shuijin Lin}
\fnmark[1]

\author[aff2]{Shaozhen Yan}
\fnmark[1]

\author[aff3,aff4]{Tuo Liu}
\fnmark[1]

\author[aff3,aff4]{Haifeng Wang}

\author[aff3,aff4]{Chunfeng Lian}

\author[aff2]{Jie Lu}
\cormark[1]
\ead{imaginglu@hotmail.com}

\author[aff1,aff4]{Jianhua Ma}
\cormark[1]
\ead{jhma@xjtu.edu.cn}

\author[aff1,aff4]{Lin Xie}
\cormark[1]
\ead{linxie@xjtu.edu.cn}

\affiliation[aff1]{organization={Key Laboratory of Biomedical Information Engineering of Ministry of Education, School of Life Science and Technology, Xi'an Jiaotong University},
                addressline={No. 28, Xianning West Road},
                city={Xi'an},
                postcode={710049},
                state={Shaanxi},
                country={China}}

\affiliation[aff2]{organization={Department of Radiology and Nuclear Medicine, Xuanwu Hospital, Capital Medical University},
                addressline={No. 45, Changchun Street, Xicheng District},
                city={Beijing},
                postcode={100053},
                country={China}}

\affiliation[aff3]{organization={School of Mathematics and Statistics, Xi'an Jiaotong University},
                addressline={No. 28, Xianning West Road},
                city={Xi'an},
                postcode={710049},
                state={Shaanxi},
                country={China}}

\affiliation[aff4]{organization={Research Center for Intelligent Medical Equipment and Devices (IMED), Xi'an Jiaotong University},
                addressline={No. 28, Xianning West Road},
                city={Xi'an},
                postcode={710049},
                state={Shaanxi},
                country={China}}

\fntext[fn1]{These authors contributed equally to this work.}

\cortext[cor1]{Corresponding authors}

\begin{abstract}
The biological definition of Alzheimer's disease (AD) relies on multi-modal neuroimaging, yet the clinical utility of positron emission tomography (PET) is limited by cost and radiation exposure, hindering early screening at preclinical or prodromal stages. While generative models offer a promising alternative by synthesizing PET from magnetic resonance imaging (MRI), achieving subject-specific precision remains a primary challenge. Here, we introduce DIReCT$++$, a Domain-Informed ReCTified flow model for synthesizing multi-tracer PET from MRI combined with fundamental clinical information. Our approach integrates a 3D rectified flow architecture to capture complex cross-modal and cross-tracer relationships with a domain-adapted vision-language model (BiomedCLIP) that provides text-guided, personalized generation using clinical scores and imaging knowledge. Extensive evaluations on multi-center datasets demonstrate that DIReCT$++$ not only produces synthetic PET images ($^{18}$F-AV-45 and $^{18}$F-FDG) of superior fidelity and generalizability but also accurately recapitulates disease-specific patterns. Crucially, combining these synthesized PET images with MRI enables precise personalized stratification of mild cognitive impairment (MCI), advancing a scalable, data-efficient tool for the early diagnosis and prognostic prediction of AD.
The source code will be released on \url{https://github.com/ladderlab-xjtu/DIReCT-PLUS}.
\end{abstract}


\maketitle

\section{Introduction}\label{sec1}
Alzheimer's disease (AD), the most prevalent neurodegenerative disorder, constitutes a mounting global health crisis. Its pathological processes begin years before clinical dementia manifests, making the prodromal stage of mild cognitive impairment (MCI) a critical window for intervention~\citep{scheltens2021alzheimer}. However, the heterogeneous nature of MCI, where only a subset of individuals progress to AD, poses a significant prognostic challenge~\citep{haller2023neuroimaging}. This challenge is underscored by the modern biological definition of AD, which is based on in vivo evidence of core pathologies, specifically amyloid-$\beta$ (A$\beta$) deposition (e.g., on $^{18}$F-AV45 PET) and neuronal injury (e.g., $^{18}$F-FDG PET hypometabolism), with structural MRI quantifying associated atrophy~\citep{ossenkoppele2022amyloid,yan2021sex,haller2023neuroimaging}. In clinical practice, however, the routine acquisition of PET is hampered by high cost, limited access, and radiation exposure, restricting this multi-modal biomarker profiling to research settings and hindering timely, widespread screening~\citep{nordberg2010use}.

Generative artificial intelligence (AI) offers a transformative pathway to bridge this clinical gap by synthesizing PET biomarkers from routinely acquired MRI~\citep{dayarathna2024deep,lee2024synthesizing}. Initial studies, primarily using generative adversarial networks (GANs)~\citep{pan2018synthesizing,pan2020spatially,apostolopoulos2022applications} and, more recently, diffusion models~\citep{xie2024synthesizing,chen2025multi,theodorou2025mri2pet}, have established the feasibility of producing visually realistic PET images. Subsequent advances have sought to improve fidelity by incorporating domain knowledge, for instance, through guidance from large vision-language models (VLMs)~\citep{wang2025unisyn,liu2025direct}.  Despite these advances, a critical limitation remains: the subject-specific precision necessary for personalized diagnosis and prognostic stratification is still lacking, which is particularly crucial for a heterogeneous condition like AD~\citep{ossenkoppele2022amyloid}. This challenges stems from the inherent nature of MRI-to-PET synthesis as an ill-posed inverse problem; neurodegenration visible on MRI is typically a downstream effect of prior molecular pathologies captured by PET~\citep{jagust2018imaging}. Consequently, individuals with similar structural MRI presentations at early disease stages can exhibit vastly different pathological burdens, leading to ambiguous mapping. Overcoming this ambiguity requires an AI synthesis framework that can comprehensively model complex cross-modal and cross-tracer associations to provide strong, patient-specific guidance.

To this end, we propose DIReCT$++$ (Domain-Informed ReCTified flow), a VLM-modulated generative model for the high-fidelity synthesis of patient-specific, multi-tracer PET images from baseline MRI and textual guidance. The DIReCT$++$ framework is built on two core innovations. First, it employs an efficient 3D rectified flow architecture to directly capture the complex, high-dimensional relationships between MRI and multi-tracer PET (specifically $^{18}$F-AV-45 and $^{18}$F-FDG), enabling concurrent and rapid generation. Second, to achieve subject-specific precision, the model is conditioned via a VLM (BiomedCLIP~\citep{zhang2025multimodal}) with application-dedicated adaptations. This domain-adapted VLM integrates fundamental patient knowledge (encoded as patient-specific clinical assessments and tracer-specific imaging domain knowledge) to guide the synthesis process, ensuring the generated PET images reflect individual pathological states rather than population averages. By reconciling an efficient generative process with precise, text-based patient conditioning, our approach directly addresses the ill-posed nature of cross-modal synthesis in heterogeneous disease populations.

We rigorously evaluate DIReCT$++$ on large-scale, multi-center datasets, including the Alzheimer's Disease Neuroimaging Initiative (ADNI)~\citep{weiner2010alzheimer} and Open Access Series of Imaging Studies (OASIS)~\citep{lamontagne2019oasis}. Comprehensive experiments demonstrate that our model generates multi-tracer PET images with superior fidelity and generalizability compared to state-of-the-art methods. Crucially, the synthesized images accurately recapitulate known disease-specific patterns of amyloid deposition and hypometabolism. When combined with the original MRI scans to train deep learning classifiers~\citep{lian2018hierarchical,huang2017densely}, the synthesized PET data enable accurate personalized diagnosis and, most importantly, the stratification of MCI patients based on their risk of progression to AD. Our work establishes a translational paradigm for precision biomarker synthesis, offering a scalable and data-efficient tool for early intervention in dementia.
\section{Materials and methods}\label{sec4}

\subsection{DIReCT$++$}
\begin{figure}
    \centering
    \includegraphics[width=1\linewidth]{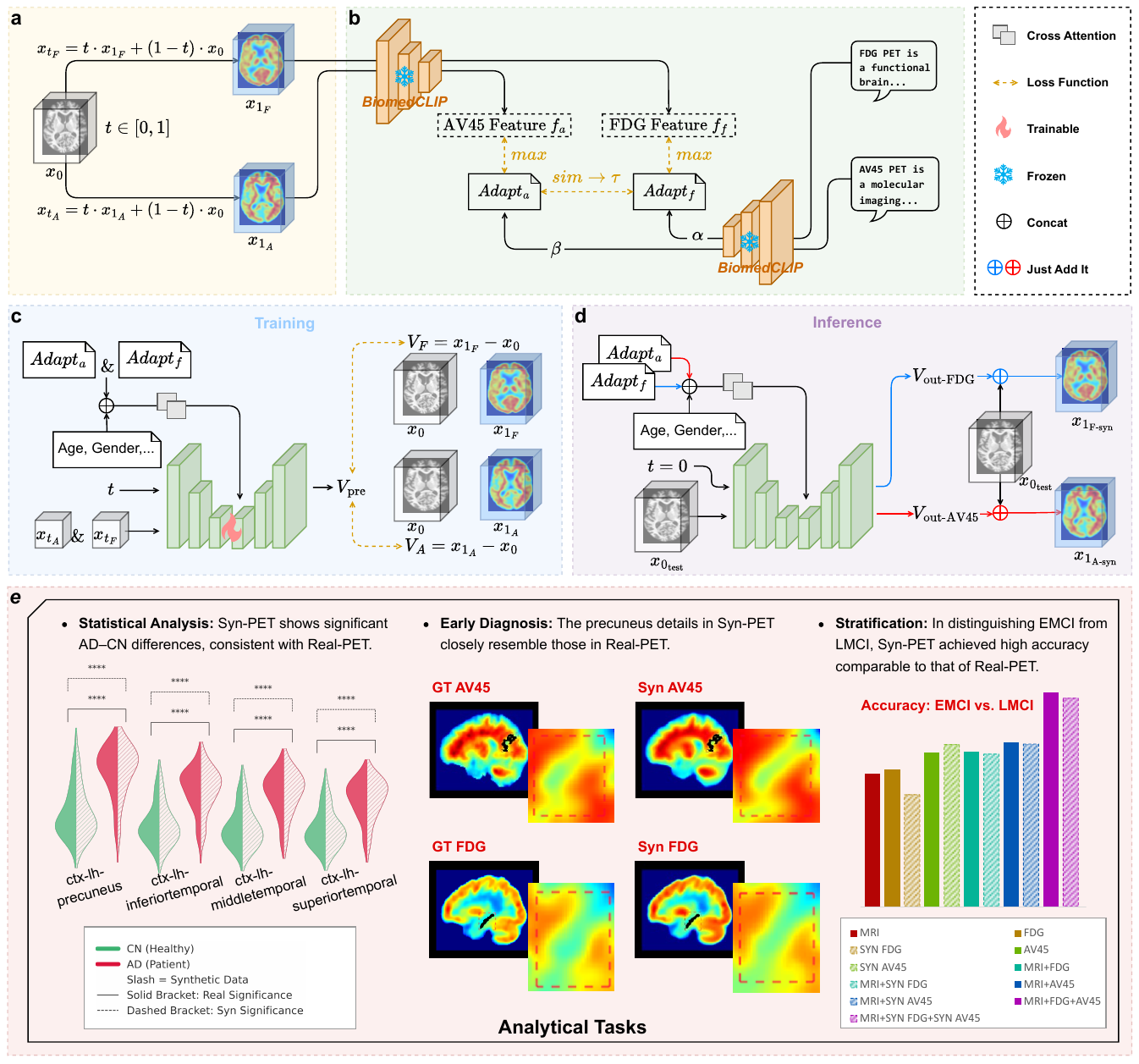}
    \caption{Overview of the DIReCT$++$ framework for multimodal PET synthesis and downstream analysis.
(a) Schematic of the rectified flow (RF) process that synthesizes dual-tracer $^{18}$F-AV-45 and $^{18}$F-FDG PET images from MRI. (b) Fine-tuning of the pre-trained BiomedCLIP model via prompt learning to encode tracer-specific and subject-level text guidance. (c) Training pipeline of the conditional 3D RF model implemented as a multi-task U-Net that predicts velocity fields for both tracers. (d) One-step inference process that simultaneously generates synthetic AV45- and FDG-PET from MRI and textual input. (e) Downstream analytical tasks, including statistical analysis, early diagnosis, and stratification, demonstrating the clinical applicability of the synthesized PET images.}
    \label{fig:1}
\end{figure}

DIReCT$++$ is a VLM-modulated rectified flow (RF) framework for synthesizing $^{18}$F-AV-45 and $^{18}$F-FDG PET images from a T1-weighted MRI scan, conditioned on text prompts derived from patient information and tracer knowledge (Fig.~\ref{fig:1}). The framework integrates a pre-trained VLM, fine-tuned from BiomedCLIP via prompt learning, with a conditional 3D rectified flow model. The VLM encodes text guidance, which is incorporated into the flow model via cross-attention to achieve subject-wise precision. The prompt learning strategy is designed to maximize intra-tracer text-image alignment while preserving inter-tracer distinctions, reflecting the complementary nature of the biomarkers. The following subsections detail the architecture and training.

\subsubsection{Domain-Adapted VLM}
To incorporate domain-specific knowledge into the generation process, we develop an adaptive guidance mechanism that leverages a frozen, pre-trained BiomedCLIP model. Specifically, we learn two \textit{affine} transformations, $Adapt_f$ and $Adapt_a$, to align the textual embeddings of FDG and AV-45 PET modalities with their respective image feature spaces. Let $c_f, c_a$ denote the BiomedCLIP textual embeddings for FDG and AV-45, respectively. 
For simplicity, we adopt lightweight affine transformations parameterized by a scalar scale and a vector bias:
\[
Adapt_f(c_f) = \alpha_1 \, c_f + \alpha_2, \quad
Adapt_a(c_a) = \beta_1 \, c_a + \beta_2,
\]
where $\alpha_1, \beta_1 $ are learnable scalar weights and $\alpha_2, \beta_2 $ are learnable bias vectors. We choose scalar scaling rather than full linear projection to minimize trainable parameters and avoid overfitting, given the limited size of medical imaging datasets. The optimal parameters are obtained by solving a constrained optimization problem:
\begin{equation}
\begin{aligned}
& \underset{Adapt_f,\, Adapt_a}{\arg \min} \quad 
   \mathcal{L}_{\text{align}} = \big(1 - \operatorname{sim}(Adapt_f(c_f), \mathbb{E}[f_f])\big) + \big(1 - \operatorname{sim}(Adapt_a(c_a), \mathbb{E}[f_a])\big) \\
& \text{subject to} \quad 
   \operatorname{sim}\big(Adapt_f(c_f), Adapt_a(c_a)\big) \geq \tau,
\end{aligned}
\label{eq:align_opt}
\end{equation}
where $\operatorname{sim}(\cdot, \cdot)$ denotes cosine similarity, and $\tau \in [0,1]$ is a predefined threshold. The constraint ensures the cross-modality guidance vectors maintain a certain level of semantic consistency.

By solving~\eqref{eq:align_opt}, we obtain the modality-specific guidance vectors $Adapt_f(c_f)$ and $Adapt_a(c_a)$, which are subsequently used to steer the rectified flow translation process in a domain-informed manner.

\clearpage

\subsubsection{Rectified Flow}
The core of DIReCT$++$ is a continuous generative dynamics framework grounded in the principle of rectified flow, designed to construct a near-optimal, domain-informed trajectory that minimizes sampling complexity while preserving the structural and semantic fidelity of medical image synthesis. Specifically, given two heterogeneous medical distributions---$\mathbf{x}_0 \sim \mathcal{M}_{\text{MRI}}$ and $\mathbf{x}_1 \sim \mathcal{P}_{\text{PET}}$ (multi-tracer)---we model the transformation between them via a learnable velocity field $V_\theta$ denotes the network parameters and $c$ encodes modality-specific contextual cues (e.g., anatomical priors or clinical metadata). This velocity field, parameterized by a U-Net architecture tailored for high-dimensional medical imagery, governs the continuous transport of samples along a deterministic path governed by an ordinary differential equation (ODE):
\begin{equation}
\frac{d\mathbf{x}_t}{dt} = V_\theta\big(\mathbf{x}_t,\, \mathbf{c},\, t\big), \quad t \in [0, 1].
\end{equation}

The modality-specific guidance embeddings $\text{Adapt}_f(c_f)$ and $\text{Adapt}_a(c_a)$ are pre-computed by optimizing Eq.~\eqref{eq:align_opt} and remain fixed throughout training and inference. These pre-computed vectors are subsequently used to steer the rectified flow translation process in a domain-informed manner. Specifically, for each subject, the composite conditioning context $\mathbf{c} $ is formed by stacking the appropriate modality guidance vector with a subject-specific clinical embedding along the token dimension:
\begin{equation}
\mathbf{c} = 
\begin{bmatrix}
\text{Adapt}_m(c_m) \\
c_{\text{demo}}
\end{bmatrix}
,
\label{eq:composite_conditioning}
\end{equation}
where $m \in \{f, a\}$ denotes the target PET tracer ($f$: FDG; $a$: AV45). Here, $\text{Adapt}_m(c_m)$ is the fixed guidance embedding obtained from the alignment optimization, and $c_{\text{demo}} $ is a frozen BiomedCLIP text embedding derived from a standardized clinical prompt encoding the individual’s demographic and cognitive profile (e.g., age, sex, weight, and neuropsychological scores).

This two-token context representation is injected into the velocity U-Net via cross-attention layers at multiple resolutions, enabling spatial features at each scale to dynamically attend to both the target imaging modality and the patient’s unique clinical attributes. Critically, no additional trainable parameters are introduced for clinical metadata encoding—all semantic information is derived from the pre-trained BiomedCLIP model—thereby enhancing generalizability and mitigating overfitting in data-limited medical imaging scenarios.

This ODE governs the transition from the initial state $\mathbf{x}_0 \in \mathcal{M}_{\text{MRI}}$ to the final state $\mathbf{x}_1 \in \mathcal{P}_{\text{PET}}$, ensuring that the velocity field $V_\theta(\mathbf{x}_t, \mathbf{c}, t)$ drives the flow along the desired path. To determine the velocity field $V_\theta$, we solve a straightforward least squares regression problem:
\begin{equation}
\min_{\theta} \int_0^1 \mathbb{E} \left\| (\mathbf{x}_1 - \mathbf{x}_0) - V_\theta(\mathbf{x}_t, \mathbf{c}, t) \right\|^2 dt, \quad \text{with} \quad \mathbf{x}_t = t\mathbf{x}_1 + (1-t)\mathbf{x}_0,
\label{3}
\end{equation}
where $\mathbf{x}_t$ represents the linear interpolation of $\mathbf{x}_0$ and $\mathbf{x}_1$. In order to train the velocity U-Net, we formulate the loss function based on Eq.~\eqref{3} as follows:
\begin{equation}
\mathcal{L}(\mathbf{x}_0, \mathbf{x}_1; \theta) = \mathbb{E}_{t,\, \mathbf{x}_0 \sim \mathcal{M},\, \mathbf{x}_1 \sim \mathcal{P}} \left\| (\mathbf{x}_1 - \mathbf{x}_0) - V_\theta(\mathbf{x}_t, \mathbf{c}, t) \right\|^2,
\end{equation}
where $t \sim \mathcal{U}(0, 1)$ is sampled from the uniform distribution.
\subsubsection{Multi-Task Learning and One-Step Generation}
To accommodate subjects for whom only one PET tracer is available, we define a unified training objective $\mathcal{L}(\theta)$ that conditionally supervises the model based on the presence of ground-truth PET data. Specifically, the total loss combines modality-specific reconstruction terms, each gated by a binary availability indicator:
\begin{equation}
\mathcal{L}(\theta) = \lambda_F \, \delta_F \, \mathcal{L}_{\text{FDG}}(\mathbf{x}_{\text{MRI}}, \mathbf{y}_f; \theta) + \lambda_A \, \delta_A \, \mathcal{L}_{\text{AV45}}(\mathbf{x}_{\text{MRI}}, \mathbf{y}_a; \theta),
\end{equation}
where $\mathbf{x}_{\text{MRI}}$ denotes the input magnetic resonance image, $\mathbf{y}_f$ and $\mathbf{y}_a$ are the corresponding ground-truth FDG and AV45 PET images (when available), and $\delta_F, \delta_A \in \{0,1\}$ indicate the presence ($1$) or absence ($0$) of each PET modality for a given subject. The hyperparameters $\lambda_F$ and $\lambda_A$ balance the relative contribution of each task during training. This formulation ensures that supervision is applied solely to available modalities, enabling robust training in heterogeneous clinical cohorts with incomplete multi-tracer imaging.
A key advantage of rectified flow is its ability to support exact one-step sampling. In DIReCT$++$, the target PET image is generated directly from the input MRI in a single forward pass, bypassing iterative refinement. 

To further enhance the accuracy of one-step generation, we incorporate a distillation step following the initial ODE-based training phase. This procedure trains the velocity network to directly predict the end-point mapping from $\mathbf{x}_0$ to $\mathbf{x}_1$, effectively collapsing the continuous flow into a single, refined inference step. The resulting one-step sampler is both faster and more precise, making it suitable for time-sensitive clinical applications. Mathematically, the generation process is expressed as:
\begin{equation}
\mathbf{x}_1 = \mathbf{x}_0 + \tilde{V}_\theta(\mathbf{x}_0, \mathbf{c}, 0),
\end{equation}
where $\mathbf{x}_0 $ denotes the subject-specific MRI input, $\mathbf{x}_1 $ the corresponding synthesized PET images, $\mathbf{c}$ the structured conditioning context encoding both anatomical priors and clinical metadata (defined in Eq.~\eqref{eq:composite_conditioning}), and $\tilde{V}_\theta$ the distilled velocity field implemented by the velocity U-Net.
\subsection{Materials and Data Preprocessing}
This study leveraged neuroimaging data from 1,857 participants enrolled in the Alzheimer’s Disease Neuroimaging Initiative (ADNI), spanning cohorts from ADNI-1 through ADNI-4. All participants underwent structural T1-weighted MRI, while subsets also had \textsuperscript{18}F-FDG-PET ($n = 1,598$) and \textsuperscript{18}F-AV45-PET ($n = 1,365$) scans available, enabling multi-tracer PET synthesis.

Participants were stratified into five clinically defined diagnostic categories to reflect the full continuum of Alzheimer’s disease pathology: Cognitively Normal (CN; $n = 577$), Alzheimer’s disease dementia (AD; $n = 317$), and three subtypes of Mild Cognitive Impairment (MCI). To account for the heterogeneity of MCI, we further subclassified individuals based on delayed recall performance on the WMS-R Logical Memory II Story A task: Early MCI (EMCI; $n = 291$) denoted a milder impairment stage, whereas Late MCI (LMCI; $n = 152$) indicated proximity to dementia conversion. The remaining MCI participants without clear progression status were grouped as MCI-unclassified (MCI-U; $n = 520$). This granular stratification facilitates a biologically grounded analysis of neuroanatomical and metabolic alterations across the AD spectrum, thereby enhancing the clinical validity of synthesized PET images.

A consistent gradient in disease severity was observed across groups. Mean Mini-Mental State Examination (MMSE) scores declined progressively from 29.40 in CN to 20.84 in AD ($p < 0.001$, trend across groups). Correspondingly, functional and neuropsychiatric burden increased with diagnostic severity, as evidenced by higher scores on the Functional Activities Questionnaire (FAQ), Geriatric Depression Scale (GDSCALE), and Neuropsychiatric Inventory–Questionnaire (NPI-Q) in AD relative to CN and MCI. Significant intergroup differences were also observed in demographic variables, including age and sex distribution (see Extended Data Table 1).

To further validate the generalizability of our model beyond ADNI, we performed an independent external evaluation using data from the Open Access Series of Imaging Studies (OASIS-3 ), a longitudinal, multi-modal neuroimaging and clinical cohort focused on normal aging and Alzheimer’s disease (available at: \url{https://www.medrxiv.org/content/10.1101/2019.12.13.19014902v1}). To ensure methodological consistency with our primary analysis, we restricted our OASIS-3 analysis to individuals with \textsuperscript{18}F-AV45 PET scans processed through the standardized Pet Unified Pipeline (PUP), which provides quantitatively reliable, cross-site standardized SUV metrics. This yielded a high-quality cohort of 419 participants. In contrast, \textsuperscript{18}F-FDG-PET data from a smaller subset ($n = 117$) were excluded due to lack of PUP processing, preserving homogeneity in image quantification. The demographic characteristics of this OASIS-3 validation cohort are detailed in Extended Data Table 2.

To ensure precise spatial correspondence between MRI and dual-modality PET images, we implemented a standardized preprocessing pipeline. T1-weighted MRI scans were first skull-stripped using FreeSurfer~\citep{Freesurfer} and subjected to histogram matching for contrast normalization. Each PET image was then rigidly registered to its corresponding MRI. Both modalities were jointly affinely normalized to the MNI152 standard space using SPM~\citep{spm}, with identical transformation matrices applied to preserve cross-modality alignment. Following spatial normalization, all volumes were cropped to a common field of view of $160 \times 192 \times 160$ voxels ($1\,\text{mm}^3$ isotropic resolution) and intensity-normalized to the $[0, 1]$ range via min–max scaling. For quantitative regional analysis, brain parenchyma was parcellated into 98 anatomical structures using SynthSeg~\citep{synthseg}.

Model evaluation was conducted under a two-fold cross-validation scheme: the cohort was randomly partitioned into two equal-sized subsets, with alternating roles as training/validation and test sets to ensure unbiased performance estimation.
\subsection{Experimental Setup and Evaluation}
\subsubsection{Competing Methods}
To rigorously benchmark the performance of DIReCT$++$, we selected six representative baseline methods spanning diverse generative paradigms, including unsupervised translation, diffusion-based synthesis, and segmentation-guided generation. All models were trained and evaluated on identical data splits, with hyperparameters optimized to their reported best configurations to ensure a fair and reproducible comparison.

\begin{enumerate}
    \item \textbf{CycleGAN}~\citep{zhu2017unpaired}: A widely adopted unsupervised image-to-image translation framework that enforces cycle-consistency to learn bidirectional mappings between unpaired domains. We include it as a representative of non-probabilistic, GAN-based cross-modality synthesis.

    \item \textbf{Swin-UNet}~\citep{Cao2021swin}: A hybrid architecture combining Swin Transformer blocks with a U-Net encoder–decoder structure, excelling in high-resolution medical image generation and fine-grained anatomical modeling.

    \item \textbf{SegGuidedDiff}~\citep{konz2024anatomically}: A state-of-the-art diffusion model that leverages segmentation maps as anatomical priors to guide PET synthesis. It serves as a high-fidelity reference for structure-preserving generative performance.

    \item \textbf{3D DDIM}~\citep{Graf2023}: A deterministic fast-sampling variant of 3D diffusion models, included to assess DIReCT$++$’s computational efficiency and sampling quality relative to accelerated diffusion approaches.

    \item \textbf{3D Rectified Flow (3D RF)}~\citep{Xing2022}: A flow-based generative model that shares DIReCT$++$’s underlying ODE framework but lacks domain-informed conditioning. This ablation model isolates the contribution of our BiomedCLIP-enhanced guidance module.

    \item \textbf{DIReCT (Single-Modality Variant)}~\citep{liu2025direct}: A simplified version of our framework that synthesizes only one PET tracer type (either \textsuperscript{18}F-FDG or \textsuperscript{18}F-AV45) per forward pass. In contrast to DIReCT$++$—which jointly generates both tracers through cross-modality interaction—this variant enables direct assessment of the benefits conferred by multi-tracer synergistic generation in terms of anatomical consistency and quantitative accuracy.
\end{enumerate}
\subsubsection{Regional Group-Wise Comparison}

To evaluate the clinical validity of synthesized PET (Syn-PET) images, we conducted a region-of-interest (ROI)-based statistical analysis focused on canonical Alzheimer’s disease (AD)-vulnerable brain regions. T1-weighted MRI scans were first parcellated into 98 anatomical ROIs using SynthSeg~\citep{synthseg}. Among these, we prioritized a set of AD-relevant regions: for metabolic assessment with \textsuperscript{18}F-FDG PET, the bilateral hippocampus, temporoparietal junction, and posterior cingulate cortex ~\citep{mueller2005alzheimers}; and for amyloid burden evaluation with \textsuperscript{18}F-AV45 PET, the precuneus, frontal lobes, and temporal lobes~\citep{wong2010in}.

The validation comprised two complementary analyses:

\begin{enumerate}
    \item \textbf{Intra-group consistency}: For each diagnostic group (AD and CN), we performed paired two-sided $t$-tests to compare the mean standardized uptake values (SUVs) between real and Syn-PET images within each key ROI. The null hypothesis was that the mean difference in SUV between real and synthesized scans is zero. A non-significant result ($p > 0.05$, corrected for multiple comparisons) would indicate high fidelity of Syn-PET to ground-truth PET at the group level.

    \item \textbf{Inter-group discriminability}: To assess whether Syn-PET preserves diagnostic signal, we conducted independent two-sided $t$-tests comparing mean SUVs between the AD and CN groups in the same ROIs—first using real PET and then repeating the analysis with Syn-PET. Preservation of statistically significant group differences (e.g., hypometabolism in FDG, elevated amyloid in AV45) in Syn-PET data would demonstrate its capacity to recapitulate clinically meaningful biomarkers.
\end{enumerate}

All statistical tests were two-tailed, and $p$-values were adjusted for multiple ROI comparisons using the false discovery rate (FDR) procedure~\citep{benjamini1995controlling}. This dual-validation framework ensures that Syn-PET not only approximates real PET quantitatively but also retains its discriminative power for early diagnostic stratification.

\subsubsection{Diagnosis and Prognostic Stratification}

To rigorously evaluate the clinical utility of synthesized PET (Syn-PET) images, we assessed their capacity to support diagnostic and prognostic classification across the AD continuum. Specifically, we trained identical DenseNet classifiers~\citep{huang2017densely} to discriminate between three clinically salient diagnostic pairs: (1)  AD versus CN individuals, a canonical diagnostic boundary; (2)  MCI versus CN, representing early disease detection; and (3) EMCI versus late MCI (LMCI), a prognostic distinction reflecting differential progression risk. 

For each classification task, models were independently trained and evaluated using the following input modalities: (i) real \textsuperscript{18}F-FDG and/or \textsuperscript{18}F-AV45 PET; (ii) Syn-PET counterparts; (iii) structural MRI alone; and (iv) multi-modal combinations, including MRI\,+\,Syn-FDG, MRI\,+\,Syn-AV45, and MRI\,+\,Syn-FDG\,+\,Syn-AV45. All models were optimized on identical training protocols and validated via 3-fold cross-validation with stratified sampling to preserve class distribution across folds. Performance was quantified using accuracy (ACC), sensitivity (SEN), and specificity (SPE).

This comparative framework enables a direct assessment of whether Syn-PET preserves the discriminative biomarker information encoded in real PET, and whether multi-tracer Syn-PET enhances early stratification beyond single-modality or MRI-only approaches. Crucially, equivalence in classification performance between real and Syn-PET would indicate that the synthesized images retain sufficient biological and pathological fidelity for use in diagnostic pipelines.
\subsubsection{Implementation Details}

All models were implemented in PyTorch (v2.6.0) and trained on NVIDIA RTX A6000 GPUs (48 GB memory per card). The Adam optimizer was used with an initial learning rate of $2.5 \times 10^{-5}$. Training was conducted for 165 epochs, with knowledge distillation applied from epoch 140 onward. Automatic Mixed Precision (AMP) was employed to accelerate training and reduce memory consumption. Each model required approximately 138 GPU hours for full convergence. A complete list of software dependencies and their versions is provided in Extended Data Table 3 for full reproducibility.

\section{Results}\label{sec2}

\subsection{The DIReCT$++$ Framework for Precision PET Synthesis}
DIReCT$++$ is a VLM-modulated rectified flow framework that synthesizes high-fidelity $^{18}$F-AV-45 and $^{18}$F-FDG PET images from a single routine MRI scan, guided by fundamental text prompts (Fig.~\ref{fig:1}a). The core of our method is a conditional 3D rectified flow (RF) model, implemented as a multi-task U-Net that predicts velocity fields (Fig.~\ref{fig:1}c). This architecture establishes a direct mapping from MRI to multi-tracer PET, enabling high-fidelity image generation in a single sampling step.

To achieve subject-wise precision, text guidance incorporating both common knowledge (e.g., general descriptions of PET tracers) and personal patient information (e.g., age, gender, clinical scores) is encoded by a pre-trained VLM and integrated into the RF model via cross-attention. We fine-tune the VLM from BiomedCLIP using a straightforward prompt learning strategy (Fig.~\ref{fig:1}b). This strategy maximizes the intra-tracer alignment between text and image embeddings while preserving the inherent inter-tracer differences, reflecting the distinct yet complementary molecular activities captured by $^{18}$F-AV-45 and $^{18}$F-FDG PET.

After training, DIReCT$++$ performs one-step inference: given an input MRI and corresponding text, it simultaneously generates the corresponding $^{18}$F-AV-45 and $^{18}$F-FDG PET images (Fig.~\ref{fig:1}d). The resulting synthetic images maintain both subject-wise and tracer-wise realism, making them directly applicable to downstream analytical tasks such as statistical analysis, early diagnosis, and stratification (Fig.~\ref{fig:1}e).

\subsection{Superior Fidelity and Generalizability of Synthesized PET}
We rigorously evaluated the multi-tracer PET synthesis performance of DIReCT$++$ against a comprehensive suite of representative methods, including generative adversarial networks (CycleGAN~\citep{zhu2017unpaired}), transformer-based architectures (Swin-UNet~\citep{Cao2021swin}), and diffusion/flow-based models (SegGuidedDif ~\citep{konz2024anatomically}, 3D DDIM~\citep{Graf2023}, 3D RF ~\citep{Xing2022}), alongside our baseline model, DIReCT~\citep{liu2025direct}, which performs mono-tracer generation without prompt adaptation. Image quality was quantified using structural similarity (SSIM), peak signal-to-noise ratio (PSNR), mean squared error (MSE), and mean absolute error (MAE) against real PET ground truth. 
Models were trained on the ADNI dataset and evaluated under two settings: internally via 3-fold cross-validation on ADNI, and externally on the independent OASIS cohort (for which only $^{18}$F-AV-45 was available).

\begin{figure}
    \centering
    \includegraphics[width=0.7\linewidth]{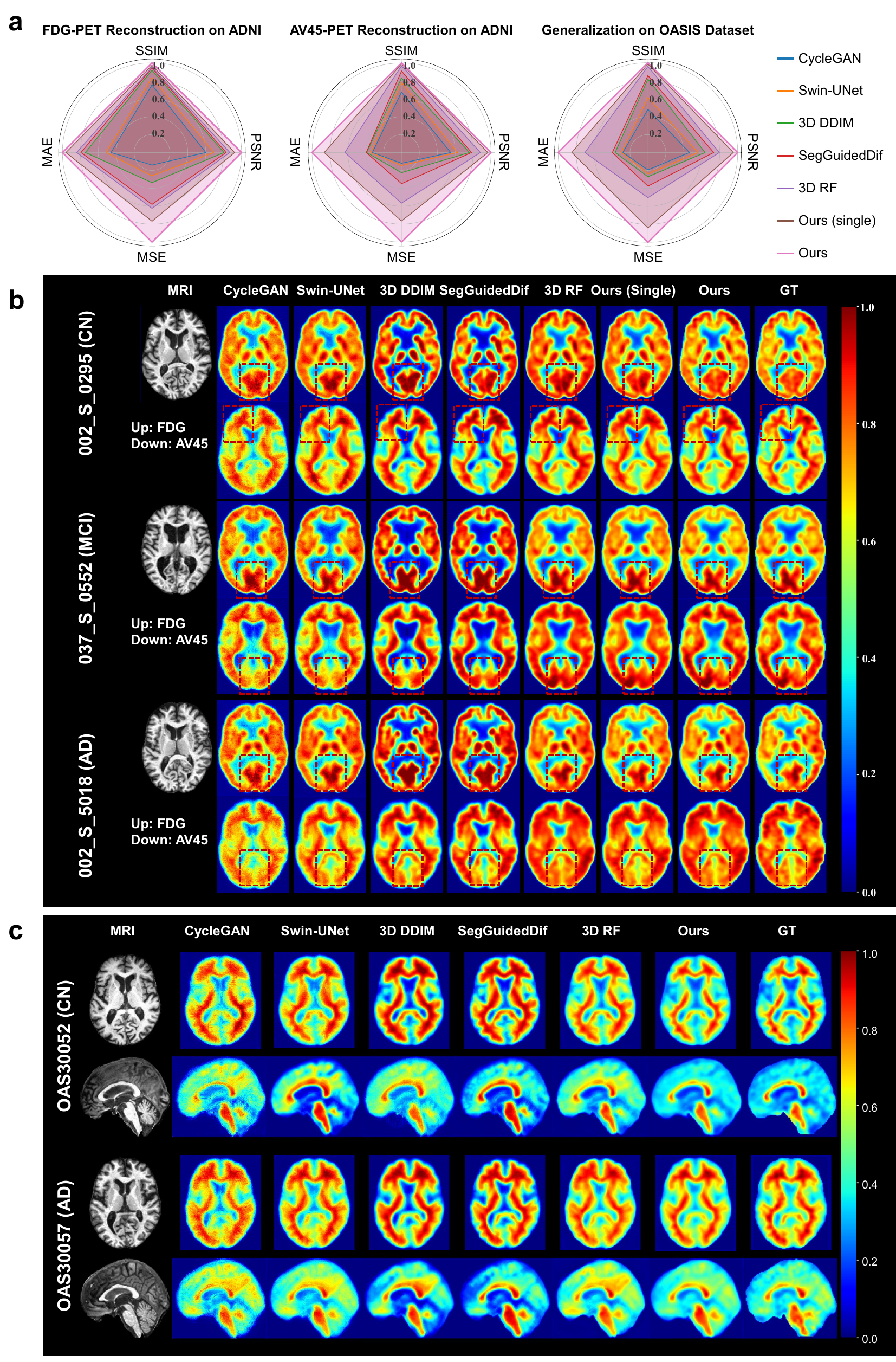}
    \caption{Comparison of PET reconstruction quality across methods and datasets.
(a) Radar plots show normalized quantitative results (SSIM, MSE, MAE, PSNR) for FDG-PET and AV45-PET reconstruction on ADNI, and cross-dataset generalization on OASIS. Notably, each metric is normalized by the best value among all methods. 
(b) Representative PET synthesis results for cognitively normal (CN, top), mild cognitive impairment (MCI, middle), and Alzheimer’s disease (AD, bottom) subjects. Columns: MRI input, competing methods, DIReCT (i.e., ours single), ours (DIReCT$++$), and real PET. Each sub-row depicts FDG (Up) and AV45 (Down) reconstruction. 
(c) Generalization to OASIS using models trained on ADNI. Rows: CN and AD subjects. Our method yields consistently higher quantitative scores and superior visual fidelity across datasets.}
    \label{fig:2}
\end{figure}

On internal evaluation, DIReCT$++$ demonstrated consistent superiority, outperforming all competing methods by substantial margins for both $^{18}$F-AV-45 and $^{18}$F-FDG synthesis (Fig.~\ref{fig:2}a). Quantitative gains were significant (e.g., $\Delta$ $+2.21$ to $+11.46$ dB in PSNR for $^{18}$F-FDG; $\Delta$ $+0.98$  to $+12.66$ dB for $^{18}$F-AV-45), with similar robust improvements observed across all metrics. This quantitative superiority was corroborated by qualitative assessment (Fig.~\ref{fig:2}b). 
\clearpage
Syntheses from DIReCT$++$ realistically captured characteristic patterns of metabolic activity and amyloid deposition across the disease spectrum, exhibiting markedly higher fidelity than competitor outputs.

External validation on the OASIS cohort confirmed that the advantages of DIReCT$++$ generalize beyond the training data (Fig.~\ref{fig:2}a). Our model sustained its leading performance, achieving a mean PSNR of $26.79$ dB (compared to $28.09$ dB on ADNI) and outperforming competitors by $1.34$ to $14.41$ dB. Visual results on OASIS cases further demonstrated excellent generalization, with DIReCT$++$ reproducing realistic tracer uptake patterns where other methods showed deviations (Fig.~\ref{fig:2}c). This robust performance across diverse datasets affirms the state-of-the-art status of DIReCT$++$, highlighting its superior fidelity, generalization capacity, and effective cross-modal and cross-tracer integration.

\subsection{Recapitulation of Disease-Specific Biomarker Patterns}
\begin{figure}
    \centering
    \includegraphics[width=1\linewidth]{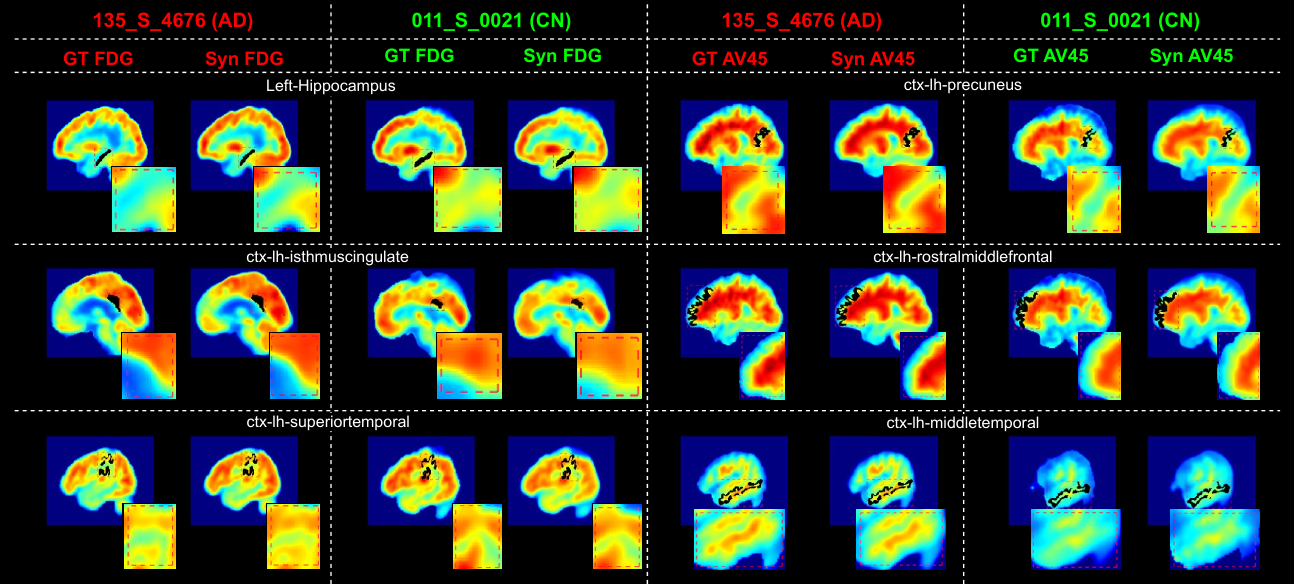}
    \caption{
Detailed regional visualization of synthetic and real PET images.
The figure displays magnified regional views of FDG-PET (left two columns) and AV45-PET (right two columns) for representative AD and CN subjects. 
For each modality and group, the first sub-column shows real PET, and the second sub-column shows synthetic PET. 
Key anatomical regions are highlighted in the bottom-right corner of each sub-column, with their names labeled at the top-center. 
Each row corresponds to a distinct brain region, demonstrating strong regional consistency between synthetic and real PET, and clear differences between diagnostic groups.
}
    \label{fig:3}
\end{figure}

To assess the clinical validity of the synthesized PET images, we parcellated both synthetic and real PET scans into anatomical regions using \texttt{mri\_synthseg} (FreeSurfer v7.4.1)~\citep{dalca2022synthseg}, a deep learning-based segmentation tool that provides high-resolution cortical and subcortical labeling consistent with the Desikan--Killiany and \textit{aseg} atlases. This approach yielded a total of 98 distinct regions of interest (ROIs), including cortical, subcortical, cerebellar, and brainstem structures. Regional standardized uptake value  (SUV) were computed by extracting the mean tracer uptake within each ROI and normalizing to the cerebellar gray matter (for \textsuperscript{18}F-AV-45 PET) or the pons (for \textsuperscript{18}F-FDG PET). Fig.~\ref{fig:3} illustrates this analysis for key ROIs in two representative ADNI subjects, demonstrating strong regional consistency between synthetic and real images, as well as clear distinctions between syntheses from different diagnostic groups.

For a comprehensive quantitative evaluation, we performed two group-wise statistical comparisons on the ADNI cohort. First, we compared synthetic and real images to validate the regional SUV precision of DIReCT$++$. Second, we compared synthetic images across diagnostic groups (cognitively normal, i.e., CN, MCI, AD) to verify their ability to capture disease-relevant patterns.

\begin{figure}
    \centering
    \includegraphics[width=0.95\linewidth]{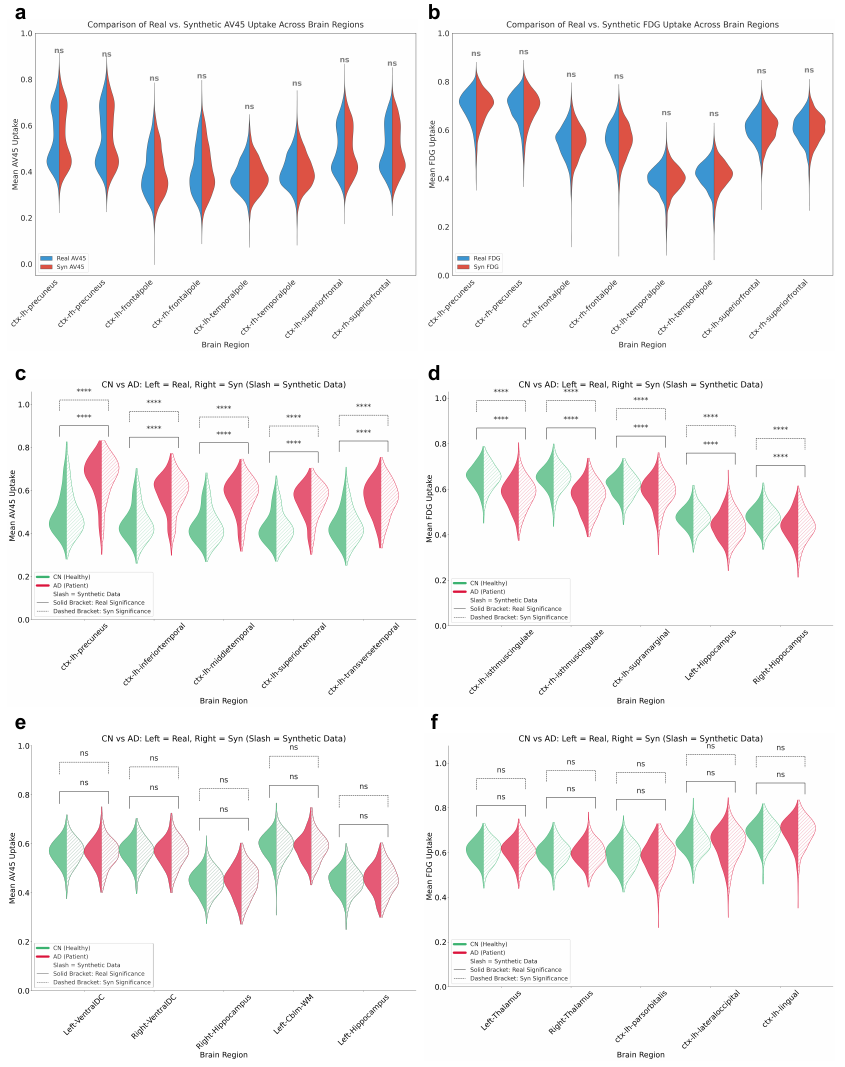}
    \caption{
Regional comparison of mean uptake between real and MRI-synthetic PET images. 
(a,b) Violin plots of regional mean uptake across cortical and subcortical regions for AV45 (a) and FDG (b), with statistical comparisons. 
(c–f) Disease-related patterns. Violin plots show mean uptake in AD (red) and CN (green). For each region, paired half-violins denote real PET (left, solid) and synthetic PET (right, dashed). Solid and dashed lines indicate significance for real and synthetic PET, respectively. 
(c,d) Disease-related regions for AV45 and FDG; (e,f) non–disease-related regions. 
Cblm-WM: Cerebellum white matter. See Extended Data for additional analyses.
}

    \label{fig:4}
\end{figure}
\clearpage
Paired-sample $t$-tests between synthetic and real images revealed no statistically significant differences in mean SUV across all principal brain regions for the CN, MCI, and AD groups ($p>0.05$; Fig.~\ref{fig:4}a, b; Extended Data Fig.1, Fig.2, and Extended Data Fig.3, Fig.4). This high fidelity was consistent for both $^{18}$F-AV-45 (Fig.~\ref{fig:4}a) and $^{18}$F-FDG (Fig.~\ref{fig:4}b), indicating that DIReCT$++$ effectively captures the regional characteristics of the ground truth.

Critically, independent-sample $t$-tests demonstrated that the significant differences between diagnostic groups observed in real PET data were precisely preserved in the synthetic images (Fig.~\ref{fig:4}c, d; Extended Data Fig.5, Fig.6). For instance, DIReCT$++$ recapitulated the statistically significant differences ($p<0.001$) between AD and CN groups in the precuneus for $^{18}$F-AV-45 (Fig.~\ref{fig:4}c) and in the hippocampus for $^{18}$F-FDG (Fig.~\ref{fig:4}d). Similarly, regional differences that were non-significant in real PET data were also consistently non-significant in the synthetic images ($p>0.05$, Fig.~\ref{fig:4}e, f). These findings confirm that DIReCT$++$ reliably preserves the essential biomarker information necessary for clinical differentiation.

\subsection{Precise Prognostic Stratification of Mild Cognitive Impairment}
To evaluate the clinical utility of the synthetic images for personalized diagnosis and prognostic stratification, we conducted classification tasks on the ADNI cohort using a 3-fold cross-validation scheme. We trained models with a DenseNet backbone~\citep{huang2017densely} to differentiate between key diagnostic pairs: (1) AD vs. CN, (2) MCI vs. CN, and (3) early MCI (EMCI) vs. late MCI (LMCI). For each task, models were trained and evaluated using various input modalities, including mono-modal (real PET, synthetic PET, or MRI alone) and multi-modal (MRI + real PET; MRI + synthetic PET) configurations. Performance was quantified by accuracy (ACC), sensitivity (SEN), specificity (SPE), and the area under the receiver operating characteristic curve (AUC).

\begin{figure}
    \centering
    \includegraphics[width=\linewidth]{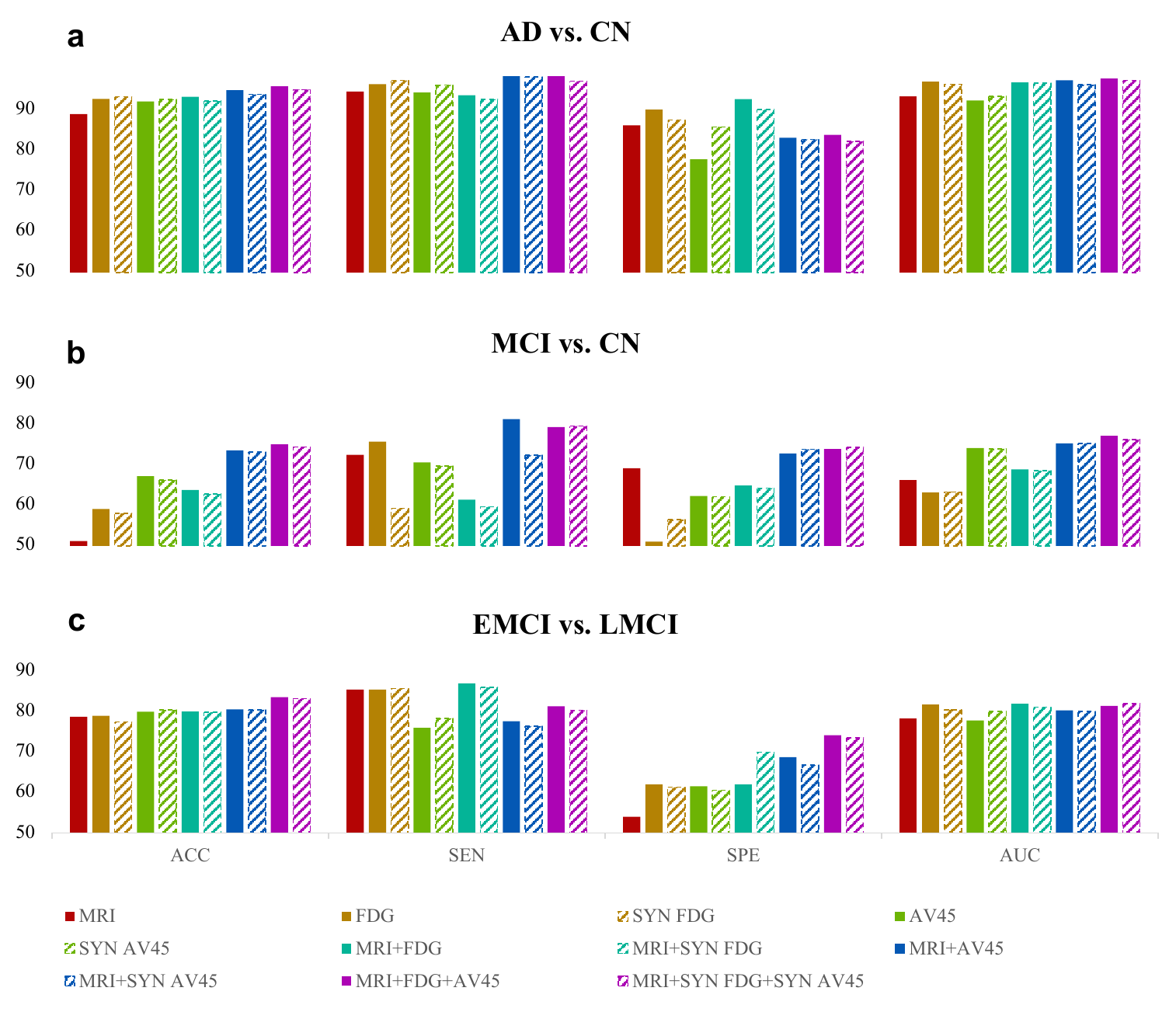}
    \caption{Diagnostic classification performance using MRI-synthetic PET images across different clinical tasks. (a–c) Bar plots illustrate the classification performance—accuracy (ACC), sensitivity (SEN), specificity (SPE), and area under the ROC curve (AUC)—of eleven input-modality combinations evaluated for three diagnostic tasks: (a) Alzheimer’s disease (AD) vs. cognitively normal (CN), (b) mild cognitive impairment (MCI) vs. CN and (c) early MCI (EMCI) vs. late MCI (LMCI). The input modalities include: MRI only, real FDG-PET, synthetic FDG-PET, real AV45-PET, synthetic AV45-PET, MRI + real FDG-PET, MRI + synthetic FDG-PET, MRI + real AV45-PET, MRI + synthetic AV45-PET, MRI + real FDG + real AV45, and MRI + synthetic FDG + synthetic AV45. For visual clarity, bars corresponding to synthetic PET inputs are shown with the same colour scheme as their real PET counterparts but with dashed (hatched) fills. The results demonstrate that MRI-synthetic PET images achieve comparable diagnostic accuracy to real PET scans, highlighting their clinical utility for downstream classification tasks.}
    \label{fig:5}
\end{figure}

The synthesized PET images from DIReCT$++$ yielded substantial performance gains that are close to real images across all tasks (Fig.~\ref{fig:5}). In the fundamental task of AD recognition (Fig.~\ref{fig:5}a), the combination of MRI with synthetic $^{18}$F-AV-45 and $^{18}$F-FDG PET achieved high performance (ACC: $93.47\%$, AUC: $95.65\%$), significantly outperforming MRI alone (ACC: $87.69\%$, AUC: $91.88\%$) and performing comparably to the combination of MRI with real multi-tracer PET (ACC: $94.27\%$, AUC: $96.10\%$).

The value of synthetic PET was even more pronounced in the challenging task of MCI diagnosis (Fig.~\ref{fig:5}b). While overall performance decreased, the improvements afforded by synthetic multi-tracer PET were more significant (e.g., SEN: $78.17\%$ for MRI+synthetic PET vs. $71.43\%$ for MRI alone), underscoring their utility for early disease detection.

Crucially, the combination of baseline MRI and synthetic multi-tracer PET also enabled accurate prognostic stratification, achieving $81.96\%$ accuracy in differentiating EMCI from LMCI, compared to $77.65\%$ using MRI alone (Fig.~\ref{fig:5}c). This result highlights the potential of DIReCT$++$ to stratify patients at the prodromal stage based on their risk of progression.

Taken together, these findings confirm the translational potential of DIReCT$++$, as its synthetic outputs serve as effective proxies for real PET biomarkers in critical downstream diagnostic and prognostic tasks.

\section{Discussion and conclusion}\label{sec3}
Our study introduces DIReCT$++$, a VLM-modulated rectified flow framework for the precise synthesis of multi-tracer PET from routine MRI under fundamental text guidance. We demonstrate that this approach not only generates images of superior fidelity and generalizability but, more importantly, yields synthetic biomarkers that accurately recapitulate disease-specific patterns and enable precise prognostic stratification of MCI. By effectively making multi-modal biomarker profiling accessible from a single MRI, DIReCT$++$ represents a scalable, AI-empowered strategy that addresses a critical bottleneck in the early diagnosis and management of Alzheimer's disease.

The superior performance of DIReCT$++$ stems from its core methodological innovations, which are designed to overcome the fundamental challenges of cross-modal medical image synthesis. First, the multi-task rectified flow (RF) architecture provides an efficient and stable framework for learning the complex, high-dimensional mapping from MRI to multi-tracer PET, enabling rapid, single-step generation of high-fidelity images. The critical innovation, however, lies in the modulation of this flow by a domain-adapted VLM. This integration directly addresses the ill-posed inverse problem of predicting molecular pathology from macrostructural anatomy. By conditioning the synthesis on patient-specific text guidance that encodes both individual clinical scores and general tracer knowledge, the model resolves the inherent ambiguity where similar MRI presentations correspond to distinct pathological states. Our results confirm this mechanism: DIReCT$++$ not only achieves superior pixel-level fidelity (Fig.~\ref{fig:2}), but also accurately recapitulates disease-specific patterns of amyloid deposition and hypometabolism (Fig.~\ref{fig:4}). Collectively, the VLM guidance and multi-task flow setting act as a powerful regularizer, steering the generation toward biologically plausible and subject-specific outcomes, thereby advancing beyond population averages to enable personalized biomarker profiling.

Beyond its technical merits, the primary significance of DIReCT$++$ lies in its potential to democratize access to critical biomarker profiling, thereby addressing a major bottleneck in the early and precise management of dementia. By generating accurate proxies for both amyloid and metabolic PET from a single routine MRI scan, our framework effectively translates the gold-standard biological definition of AD into a scalable tool for clinical practice. The clinical viability of this approach is strongly evidenced by our results on personalized MCI diagnosis and stratification (Fig.~\ref{fig:5}b, c), where the combination of MRI and synthetic multi-tracer PET achieved performance comparable to the resource-intensive ideal of MRI with real PET, significantly outperforming models using MRI or real PET alone. 

This capability has profound implications. It could enable widespread screening of individuals at the preclinical or prodromal stages in primary care settings, identifying those with underlying AD pathology who are most likely to benefit from disease-modifying therapies. Furthermore, the ability to serially generate synthetic PET biomarkers from longitudinal MRI without additional radiation exposure opens new avenues for cost-effective monitoring of disease progression and treatment response. DIReCT$++$ thus moves us closer to a future where multi-modal diagnostic precision is accessible as part of the standard clinical workflow, unconstrained by current resource limitations.

When contextualized within the field of medical image synthesis, the advancements of DIReCT$++$ represent a qualitative leap beyond prior approaches. While existing generative adversarial networks and diffusion models have demonstrated the feasibility of MRI-to-PET translation, they have primarily focused on achieving visual realism. DIReCT$++$ moves beyond this by achieving precise generation of subject-specific biomarkers that are clinically actionable. This is a direct consequence of its VLM-modulated architecture, which incorporates domain knowledge as a critical regularizer, a feature largely absent in preceding purely, data-driven approaches. The core principle of using a domain-adapted foundation model to guide an efficient generative process establishes a new paradigm that is extendible well beyond the synthesis of $^{18}$F-AV-45 and $^{18}$F-FDG PET. By prioritizing clinically grounded precision over mere voxel-level fidelity, our work points the way toward a new generation of AI models that are truly co-designed with clinical application in mind.

Several limitations of this study present clear avenues for future work. First, the current text guidance relies on fundamental patient information and imaging knowledge. Incorporating additional clinically accessible data, such as blood biomarkers~\citep{hansson2023blood}, could further enhance subject-level precision, necessitating the development of targeted cross-modal pretraining or prompt-learning strategies. Second, while DIReCT$++$ synthesizes $^{18}$F-AV-45 and $^{18}$F-FDG, extending the framework to include tau-PET imaging would provide a more comprehensive biomarker profile for sensitive screening at the preclinical stage. Third, to maximize clinical utility, future iterations could integrate downstream task guidance (e.g., diagnosis or prognosis labels) directly into the generative model, enabling controllable synthesis~\citep{wang2025flexibly,chang2025controllable} tailored for specific clinical decisions. Finally, the evaluation was conducted on well-characterized but limited research cohorts (ADNI, OASIS). Prospective validation on more diverse, real-world clinical populations (including data from thick-slice or low-field MRI~\citep{sorby2024portable}) is essential to fully establish the generalizability and robustness of the framework.

In conclusion, we have presented DIReCT$++$, a framework that synergizes rectified flow with domain-adapted vision-language modeling to achieve precise synthesis of multi-tracer PET from routine MRI. By generating clinically actionable biomarkers that enable accurate diagnosis and prognostic stratification of MCI, our work effectively bridges the gap between the biological definition of AD and its clinical application. DIReCT$++$ represents a significant step toward scalable, accessible, and precise tools for early intervention in neurodegenerative disorders.

\printcredits

\section*{Conflict of interest}
The author declares that they have no conflict of interest.

\section*{Acknowledgments}
The authors acknowledge the funding of the the National Natural Science Foundation of China (No.~T2522028 and 12326616), Natural Science Basic Research Program of Shaanxi (No.~2024JC-TBZC-09), and Shaanxi Provincial Key Industrial Innovation Chain Project (No.~2024SF-ZDCYL-02-10).

\section*{Author contributions}
S.L., S.Y., and T.L. contributed equally to this work. 
S.L. and T.L. conceived the methodology, implemented the core algorithms, and conducted the main experiments. 
S.Y. contributed to data curation, preprocessing, and experimental implementation. 
H.W. contributed to model evaluation, validation experiments, and statistical analysis. 
C.L. contributed to study design, project supervision, and interpretation of experimental results. 
J.L. provided clinical guidance and contributed to the interpretation of neuroimaging findings. 
J.M. and L.X. provided methodological guidance, technical supervision, and critical revision of the manuscript. 
S.L. and T.L. drafted the manuscript. 
All authors reviewed, edited, and approved the final manuscript. 
\clearpage
\bibliographystyle{unsrt}
\bibliography{sn-bibliography}

\end{document}